\documentclass[runningheads]{llncs}

\usepackage[T1]{fontenc}
\usepackage{graphicx}
\usepackage{multirow}
\usepackage{float}
\usepackage{amsmath}
\usepackage{amssymb}
\usepackage{color}

\title{ViMo: Generating Motions from Casual Videos}

\author{Liangdong Qiu\inst{1} \and
Chengxing Yu\inst{2} \and
Yanran Li \and
Zhao Wang\inst{3} \and
Haibin Huang\inst{4} \and
Chongyang Ma\inst{5} \and
Di Zhang\inst{4} \and
Pengfei Wan\inst{4} \and
Xiaoguang Han\textsuperscript{\dag}\inst{1}}

\institute{
\inst{} Chinese University of Hong Kong (Shenzhen) \and
\inst{} University of Electronic Science and Technology of China \and
\inst{} Zhejiang University \and
\inst{} Kuaishou Technology \and 
\inst{} ByteDance 
}

\begin{document}

\authorrunning{ViMo: Generating Motions from Casual Videos}

\maketitle       

\begin{abstract}
Although humans have the innate ability to imagine multiple possible actions from videos, it remains an extraordinary challenge for computers due to the intricate camera movements and montages. Most existing motion generation methods predominantly rely on manually collected motion datasets, usually tediously sourced from motion capture (Mocap) systems or Multi-View cameras, unavoidably resulting in a limited size that severely undermines their generalizability. Inspired by recent advance of diffusion models, we probe a simple and effective way to capture motions from videos and propose a novel Video-to-Motion-Generation framework (ViMo) which could leverage the immense trove of untapped video content to produce abundant and diverse 3D human motions. Distinct from prior work, our videos could be more causal, including complicated camera movements and occlusions. Striking experimental results demonstrate the proposed model could generate natural motions even for videos where rapid movements, varying perspectives, or frequent occlusions might exist. We also show this work could enable three important downstream applications, such as generating dancing motions according to arbitrary music and source video style. Extensive experimental results prove that our model offers an effective and scalable way to generate diversity and realistic motions. Code and demos will be public soon.
\keywords{Video-Motion Generation \and Capture from Videos \and Motion Diffusion Models}
\end{abstract}
\renewcommand{\thefootnote}{}
\footnotetext{\textsuperscript{\dag} Corresponding author.}

\begin{figure}
  \centering
  \includegraphics[width=0.99\linewidth]{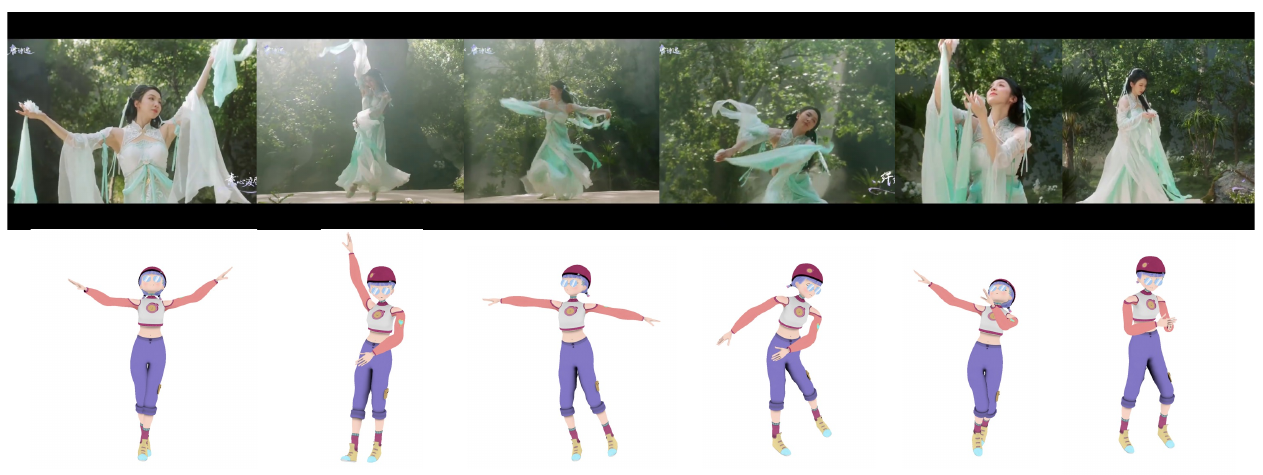}
  \caption{An example of \textbf{Casual Video.} This is a game advertisement video. The camera's perspective and distance from the actor constantly change, with some joints being obscured. The character at the bottom is the 3D motion clips generated by our method, ViMo, which can extract comparable 3D motions from casual videos. Traditional human pose reconstruction methods often fails to obtain plausible motions on these casual videos where the cameras are complex and occlusion of characters exists from beginning to end.}
  \label{fig:teaser}
  \vspace{-1mm}
\end{figure}

\section{Introduction}

Realistic 3D human motions play an important role as the `soul' of virtual characters since they can make them feel alive – it enables the static character to jump, fight and dance~\cite{aberman2020skeleton}. With the rapid proliferation of portable devices and the metaverse, the demands on virtual character creation are rising at an explosive rate. In the traditional animation and film industry, the motions are created manually in a tedious way either by expensive motion capture systems~\cite{menolotto2020motion} or by professional animators, therefore usually resulting in a very time-consuming and laborious process. Motivated by this problem, data-driven motion generation methods~\cite{tseng2023edge,raab2022modi,tevet2022human,alexanderson2023listen,huang2023diffusion,zhang2022motiondiffuse} have been widely studied recently since they are super efficient and economical~\cite{zhang2023magicavatar}.

Although promising progress in motion generations has been achieved, the diversity and flexibility of the synthesised motions are still far behind the expectations of the users in practical applications. Most existing motion generation approaches adopt the diffusion-based, VAE or GAN-based framework and take condition signals which are input parameters or data types that guide the motion generation process, such as Gaussian noise~\cite{yan2019convolutional,raab2022modi}, text~\cite{tevet2022motionclip,tevet2022human,zhang2022motiondiffuse}, audio~\cite{liu2022audio,alexanderson2023listen}, music~\cite{aristidou2021rhythm,sun2020deepdance,li2022danceformer} or scene contexts~\cite{wang2022humanise,huang2023diffusion}. The effectiveness of these models is contingent upon the availability of matched accurate examples (ground-truth pairs) of both the input conditions (like audio or text) and the corresponding motions. Consequently, this way narrows the synthesised motions only to specific domains and finite categories provided in the training dataset. For example, the state-of-the-artwork EDGE~\cite{tseng2023edge} generated dancing movements are all limited to the 10 categories of street dancing provided in the training dataset AIST++~\cite{li2021ai}. The existing action-conditioned motion generations~\cite{degardin2022generative,maheshwari2022mugl,petrovich2021action,guo2020action2motion} are only able to generate motions in the same label of the input in the training dataset as well. 

The motion datasets hamper these generation methods from scaling up. This is because they are typically captured using professional Mocap Systems or Multi-View cameras which are time-consuming and costly. 
In contrast, video resources are incredibly extensive and widely available, offering a broader range of diverse and interesting human actions. 
However, this topic remains very under-exploited due to the significant complexity and variability of video content. 
There are some existing approaches~\cite{zhu2023motionbert,yuan2022glamr,ye2023decoupling,pavlakos2022human,pavllo20193d,cai2023smpler} explored to leverage the 3D pose estimation methods to reconstruct the 3D human motions precisely from video resources, therefore generating more diverse types of motions. For example, The state-of-the-artwork MotionBERT~\cite{zhu2023motionbert} can estimate high-fidelity 3D joint positions and perform natural human movements similar to the videos. 
However, their models are mostly only tested on videos with specific conditions --the camera position could be known accurately. When the videos include complex camera movements or montage editing techniques (Fig~\ref{fig:teaser}), their performance could degrade drastically and fall into collapse. However, most online videos that we denote as \textbf{Casual videos} usually unavoidably contain complex camera movements, which are out of their assumption. 

Reconstructing 3D motions from these casual videos will be incredibly complicated, as estimating the complex camera angles and locations is overly intricate and sometimes not feasible.
Without accurate camera positions, the regression loss employed in the aforementioned methods will become completely dysfunctional and frame-wise poses are hard to string together as a smooth motion. Therefore, learning 3D motions from casual videos is a desirable goal but remains a formidable challenge due to the intricate camera issues. 

In contrast, humans can easily comprehend 3D actions from casual videos. To capture the essence of actions in videos, the 3D motions just need to be similar in poses and rhythms rather than the high precision of joint locations. Inspired by that, we pose a new video-to-motion generation question -- \textit{could the computer generate multiple possible 3D motions of the new types according to the actions provided in the casual videos which have very similar pose shapes and rhythms? } 

To achieve this goal, we propose a novel video-to-motion generation model by taking videos as conditions to generate various motions and showcase its great potential in enabling three important downstream applications. Specifically, we explore a diffusion-based framework which takes the 2D poses from multiple different views as the input data to generate 3D motions to see how far it could work on video-to-motion generation tasks. With this design, our model can create a sequence of 3D motion without explicitly estimating the camera positions. 
We conducted extensive experiments to validate the effectiveness of our approach. The results reveal that this simple approach could effectively generate unlimited complex and realistic 3D motions which are contextually consistent with the input casual videos even when they contain complex camera movements and occlusions. Notably, we show that our work could enable three interesting downstream applications. (1) Firstly, it could be an efficient way to produce large-scale motion datasets. We collected a large-scale Chinese classic dancing dataset and built up a 3D dancing motion dataset that included 750 motions. This new dataset could be used to facilitate data-driven tasks such as music-to-dancing generation, motion recognition and prediction. (2) Secondly, our model could power up few-shot stylization in the dancing generation. Specifically, we could leverage the music-to-motion generation pipeline to generate a specific style of dance according to the pose style offered by a few video examples. (3) Thirdly, we could form a new video-guided motion completion and editing task which could fill in the missing part of the motion by the similar content defined by any source videos. 

In summary, we contribute in the following aspects to encourage future research towards this goal:
(1) We propose a novel video-to-motion-generation model to leverage the abundant resources of the actions in videos to enhance the diversity and realism of 3D motion generation. By posing this new question, we reveal that the diversity and flexibility of motion generation could be greatly extended by a simple but effective approach. 
(2) A dancing specified 3D motion dataset is constructed with the proposed ViMo model, which contains 52 Chinese classic dancing videos together with 750 generated 3D dancing motions. This demontrate the effectiveness and advantages of video-to-motion-generation strategy. We hope suck kind of video-to-motion dataset could facilitate a lot of motion-related tasks.  
%
(3) We demonstrate that the proposed method could be impressively activated in three real-life applications which illustrate its astonishing ability to generate realistic and aesthetical human motions.

\section{Related Work}

\subsection{Human Motion Generation}
%
Existing human motion generation proposes generative models which take conditional signals as guidance in order to consistently generate motions. These methods could be categorized based on different types of conditional signals, including Gaussian noise, seed motion, text, music, audio and scene contexts. 

Action-conditioned motion generation sounds close to our work but they only take seed motions or an action label~\cite{degardin2022generative,maheshwari2022mugl,petrovich2021action,guo2020action2motion} to generate the motions in the same category of the reference action. In contrast, we could generate motions in an unseen category. Several work~\cite{yan2019convolutional,raab2022modi} have learned motion manifolds in which the user could sample the Gaussian noise as input to obtain new motions. However, their results on the manifold are mostly locomotion. Text-based motion generation~\cite{tevet2022motionclip,tevet2022human,zhang2022motiondiffuse} has attracted increasing attention recently due to the advanced natural language processing technologies. Such methods aim to learn an alignment embedding of the motion sequences and text description. Another stream of work generates hand gestures or dancing movements based on the audio data such as speech~\cite{zhi2023livelyspeaker,liu2022audio,yi2023generating} and music~\cite{aristidou2021rhythm,sun2020deepdance,li2022danceformer}. Scene-guided motion generation~\cite{wang2022humanise,wang2021scene,huang2023diffusion} is highly goal-oriented and requires the generated motion to be physically reasonable and interactive with the objects in the scene environment such as indoor activities. Most of the aforementioned approaches are only able to synthesise motion categories which appeared in the training dataset before. Very few works such as MotionCLIP~\cite{tevet2022motionclip} and AvataCLIP~\cite{hong2022avatarclip} could leverage the CLIP loss to generate reasonable motion sequences from unseen categories with the text labels. However, there are unlimited motions like Kongfu, dancing, and sports that cannot be aligned with any category or precise text labels. 


\subsection{Motion Reconstruction from Video}


The existing 2D to 3D pose estimation can be broadly categorized into two ways. The first type of methods~\cite{moon2019camera,sun2018integral,zhou2019hemlets} employ a convolutional neural network learning the image features from single image and regressing the 3D joints coordinates directly. The second type of methods~\cite{ci2023gfpose,martinez2017simple,cheng20203d,cai2019exploiting,wan2021encoder,ci2019optimizing,li2022mhformer,shan2022p,zhang2022mixste,zheng20213d,zhu2023motionbert,yuan2022glamr,ye2023decoupling,pavlakos2022human,pavllo20193d,cai2023smpler} extract the 2D pose from the images then proposes diverse deep learning models to lift the 2D pose into 3D space, which is current most common way. State-of-the-art work MotionBERT~\cite{zhu2023motionbert} has provided a typical approach for 3D motion reconstruction by utilising the 3D pose estimation methods. They have trained a Dual-stream Spatio-temporal Transformer framework on a 3D motion regression task. Despite these deterministic methods, a diffusion-based 3D pose estimation framework~\cite{holmquist2023diffpose} is proposed to generate stochastic multiple pose estimation results. Recently, SMPLer-X~\cite{cai2023smpler} has achieved the best performance on 3D pose estimation by scaling up the training data to be $4.5$ million level. 

Most of the aforementioned approaches achieve promising results by assuming the camera parameters are almost fixed or could be easily estimated accurately. For example, MotionBERT~\cite{zhu2023motionbert} and GLAMR~\cite{yuan2022glamr} present very accurate motion reconstruction results when the camera is estimated correctly but the regression becomes completely infeasible when the camera is too complex. Furthermore, it remains a great challenge to compose the single skeletons together as a smooth and complete 3D motion even if they could show very accurate 3D estimation results on a single image, especially when the changes in camera position and orientation are significant.


\section{Methodology}

\begin{figure}[t]
  \centering
  \includegraphics[width=1.03\linewidth]{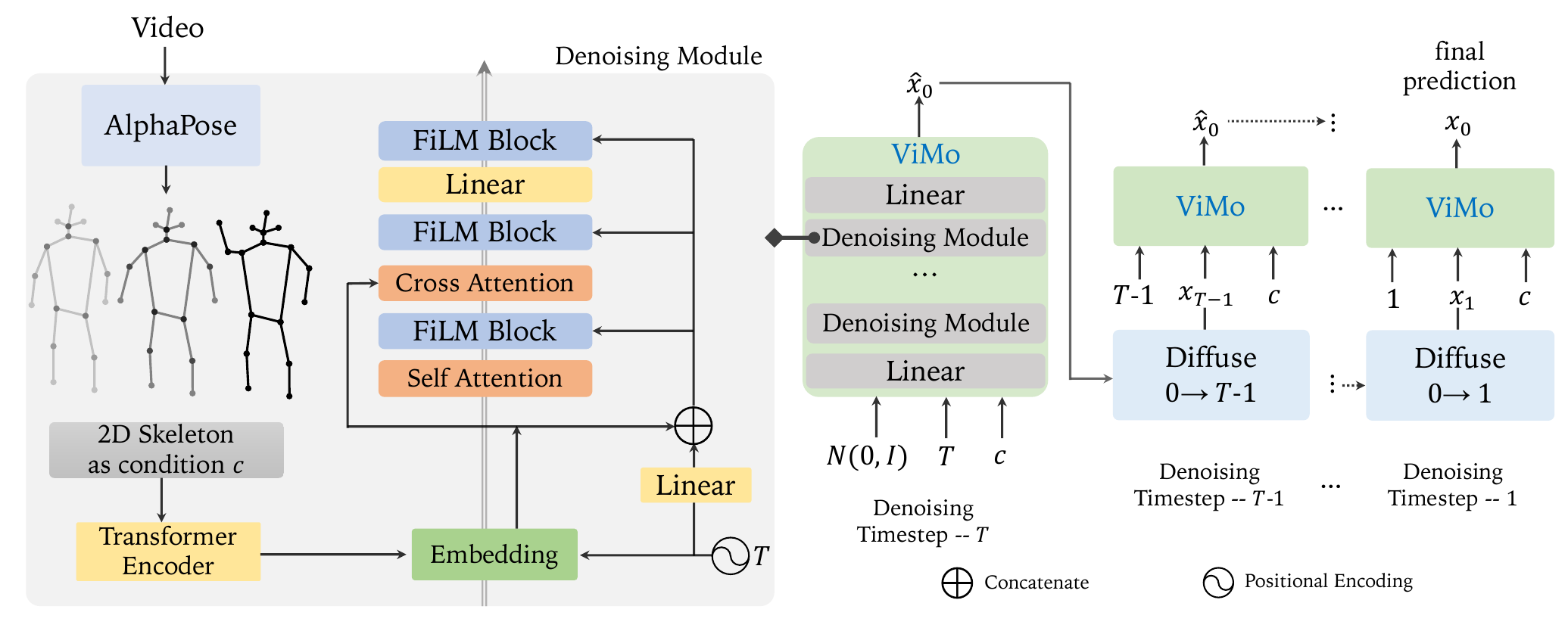}
  \caption{Diagram of our ViMo \textbf{pipeline}. ViMo takes a sequence of 2D poses as input conditional signal $c$. Then it will process a denoise process to obtain a 3D motion sequence from time $t = T$ to $t = 0$. Note that the motion itself is the prediction at each denoising step. One of the advantages is the diffusion process performs robust on the casual videos and could generate corresponding 3D motions without estimating the precise camera positions.}
  \vspace{-2mm}
\label{fig:pipeline}
\end{figure}

\begin{figure}[!h]
  \centering
  \includegraphics[width=0.88\textwidth]{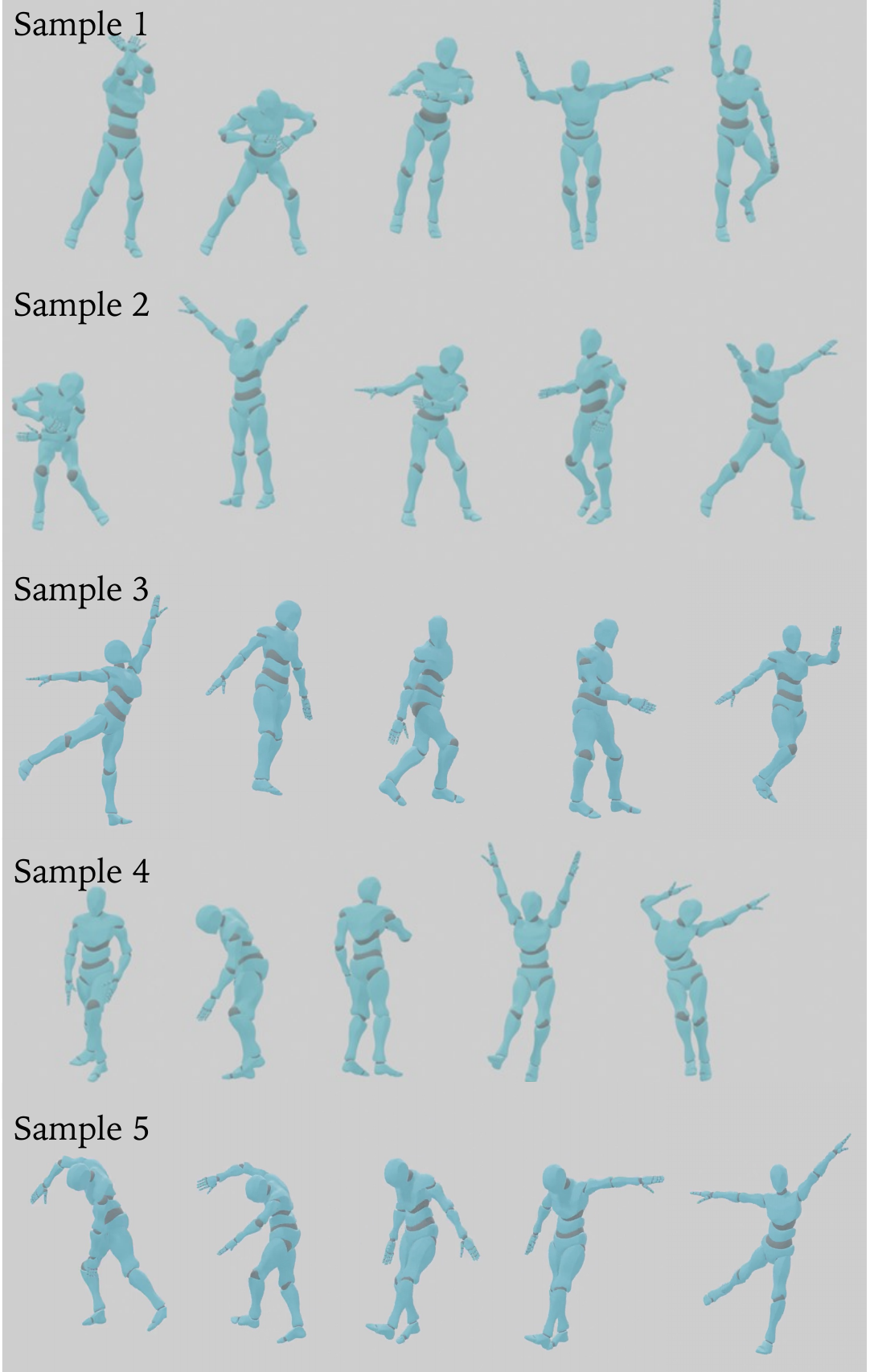}
  \caption{Constructed Chinese classic dance dataset. ViMo can help to build high-quality motion data and utilize these data to benefit downstream tasks as our other applications illustrate.}
  \vspace{-2mm}
  \label{fig:supply_chinese_dance}
\end{figure}

\begin{figure}[!h]
  \centering
  \includegraphics[width=0.98\textwidth]{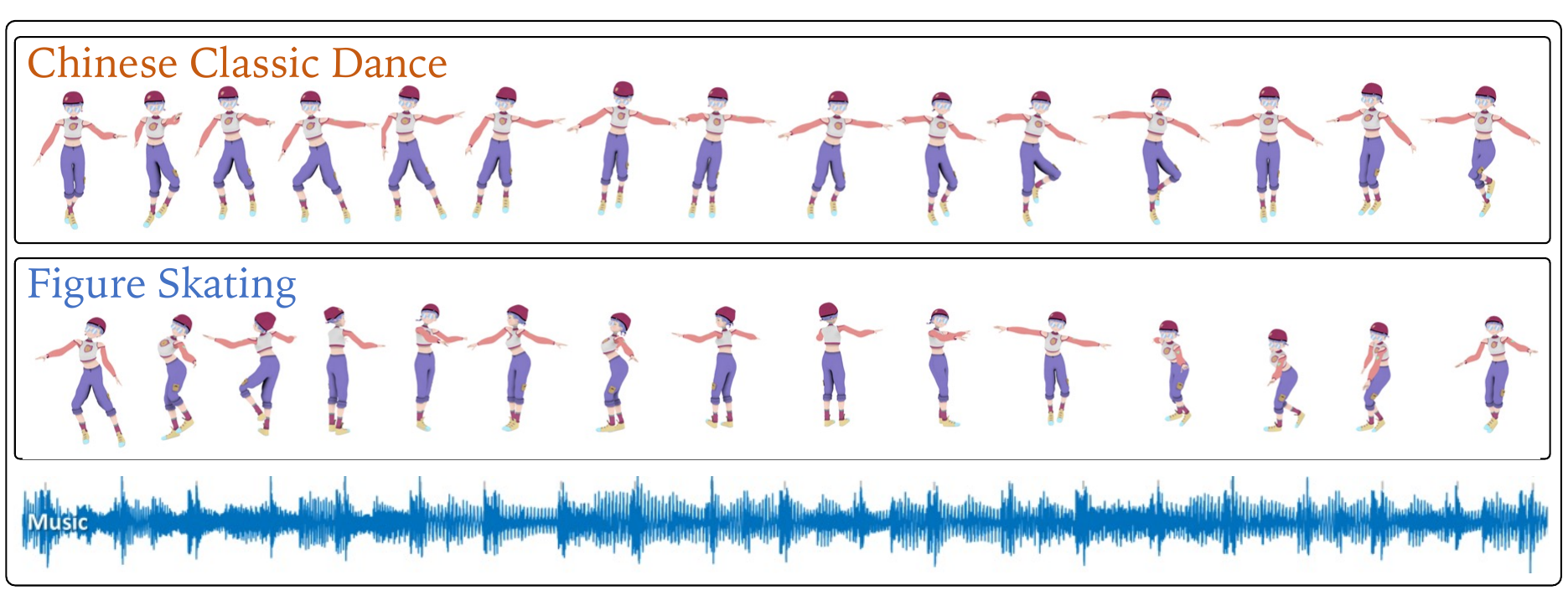}
  \caption{\textbf{Dance Motion Stylization} given arbitrary music. Given a few references videos, our ViMo can conveniently extract motions and feed these data to music-to-dance models to learn motions in the corresponding style. This enables generate one style motion, kungfu style for instance, from a totally different kind of music such as pop music.}
  \label{fig:dance}
  \vspace{-3mm}
\end{figure}

\subsection{Overview}
Given a video in the wild, which records a single person performing a certain action under various views and editing, video-to-motion generation aims to generate multiple plausible 3D human motions. The result motions are proposed to be semantically aligned to the video content in terms of pose shape and rhythm.  

To address this issue, a diffusion-based framework ViMo is elaborated to conduct the video-to-motion generation.
In the following paragraph, we further present the details of how to incorporate ViMo in three crucial downstream applications. Once the bridge of video content to 3D motion has been built, efficient and semantic control methods could be offered to the user therefore facilitating a more interactive experience in the motion applications. 

\subsection{Problem Definition}
To simplify the scenario, the input casual video only contains a single person is considered. Taking existing pose extractors such as~\cite{fang2022alphapose,chen2020fall}, the basic 2D pose estimation results could be obtained through the image features. The 2D pose sequences are denoted as $p \in \mathbb{R}^{S \times J_{2d} \times C_{\text {in}}}$. In this work, the input sequence length is set as $S=150$ and each pose in the sequences has a skeleton with 17 joints which could be written as $J_{2d} = 17, C_{\text {in}} = 3$. 

Note the input 2D skeleton follows the standard COCO format~\cite{lin2014microsoft} along with the predicted coordinates and confidence. The missing joints and low-confidence joints will be filtered out in the proposed pipeline. Our objective is to generate 3D motion $\mathbf{M}$ as a sequence of joints rotations $R \in \mathbb{R}^{J_{3d} \times C}$ in 6D rotation space, foot contact labels $\mathbb{R}^4$ and root position $\mathbb{R}^3$ where $J_{3d} = 24, C = 6$.

\subsection{ViMo framework}

\subsubsection{Diffusion Models.}
Inspired by recent successes in image generation and natural language processing tasks~\cite{saharia2022palette,ruiz2023dreambooth,ceylan2023pix2video,xu2023versatile,zhang2023adding,li2022diffusion}, a transformer based diffusion model is leveraged to align the synthesis motion to the 2D pose sequences. Considering the generation as a Markov noising Process following~\cite{ho2020denoising}, the 2D pose sequence $p$ would join the noising process to finally produce a 3D motion  $m \in \mathbb{R}^{S \times (24 \times 6+4+3)}$ corresponding to the conditional signal.

A typical diffusion process is employed as an initial step. The noising process satisfies
\begin{equation}
q\left(m_t^{1: S} \mid m_{t-1}^{1: S}\right)=\mathcal{N}\left(\sqrt{\alpha_t} m_{t-1}^{1: S},\left(1-\alpha_t\right) I\right),
\end{equation}
with latent $\left\{m_t^{1: S}\right\}_{t=0}^T$ which is assumed as Gaussian distribution and $x_0^{1: S}$ is sampled from original data $m \sim q\left(\mathbf{M} \mid p\right) $. 
$\alpha_t \in(0,1)$ is the constant parameter at noising step $t \in [0, 1, ..., T]$. 

The reversed denoising process is modelled with a neural network starting from the Gaussian distribution $m_T^{1: S} \sim \mathcal{N}(0, I)$ with simple objective~\cite{ho2020denoising} 

\begin{equation}
\hat{m}_0=D\left(m_t, t, p\right),
\end{equation}

\begin{equation}
\mathcal{L}_{\text {simple }}=E_{m \sim q\left(\mathbf{M} \mid p\right), t \sim[1, N]}\left[\left\|m_0-D\left(m_t, t, c\right)\right\|_2^2\right],
\end{equation}
where $\hat{m}_0$ obeys the original data distribution $q\left(\mathbf{M} \mid p\right)$. It is used for computing the posterior distribution $q\left(m_{t-1} \mid m_t, m_0, p\right)$ at denoising step $t$~\cite{ramesh2022hierarchical}. 
It needs to point out that result motions are directly predicted rather than reparameterization of vanilla epsilon. 

\subsubsection{Network Architecture.}
The schematic of the proposed network architecture and denoising steps are described in Fig~\ref{fig:pipeline}. Given an input $m_{t}$, three denoising modules would gradually recover the data by reducing the noise from time step $t$ to $t-1$. Each module consists of basic attention blocks and MLP layers. The first self-attention block is designed to leverage the information of the input $m_t$. During the diffusion process, the time step and 2d poses $p$ feature are incorporated together by FiLM blocks~\cite{perez2018film}. After that, the fused feature vector is sent into cross-attention and MLP layers to enhance the capability of the model. Note that, sampling processing is employed during training in a classifier-free manner and Classifier-free diffusion guidance. The resulting estimate is composed of 25\% unconditioned guidance and 75\% conditioned guidance, balancing between diversity and fidelity by allowing a 25\% probability to drop off the guidance.

\subsubsection{Loss Functions.} The generated motions are required to be continuous, smooth and physically reasonable for practical use. Therefore, existing work has proposed a set of loss functions in the typical generation task to strengthen the quality of the produced motion. We follow the standard approach and take three of the most commonly used loss functions as auxiliaries to enhance the performance. The formulations of the three commonly used losses, which are 3D positions loss, velocities loss and foot contact loss respectively, are listed in the following:

\begin{equation}
\mathcal{L}_{\text {joints}}=\frac{1}{S} \sum_{i=1}^S\left\|F K\left(R_0^i\right)-F K\left(\hat{R}_0^i\right)\right\|_2^2,
\end{equation}

\begin{equation}
\mathcal{L}_{\mathrm{vel}}=\frac{1}{S-1} \sum_{i=1}^{S-1}\left\|\left(R_0^{i+1}-R_0^i\right)-\left(\hat{R}_0^{i+1}-\hat{R}_0^i\right)\right\|_2^2,
\end{equation}

\begin{equation}
\mathcal{L}_{\text {foot}}=\frac{1}{S-1} \sum_{i=1}^{S-1}\left\|\left(F K\left(\hat{R}_0^{i+1}\right)-F K\left(\hat{R}_0^i\right)\right) \cdot \hat{f}_i\right\|_2^2.
\end{equation}
The $F K(\cdot)$ denotes the forward kinematic function converting joint rotations into joint positions. Note that we use the model’s own prediction $\hat{f}_i$ of the binary foot contact label’s following~\cite{tseng2023edge}. The overall training loss is the adaptive weighted sum of the simple loss and the auxiliary loss functions:
\begin{equation}
    \mathcal{L} = \underbrace{\mathcal{L}_{simple}}_{diffusion} + \underbrace{\lambda_1 \mathcal{L}_{joints} + \lambda_2 \mathcal{L}_{vel} + \lambda_3\mathcal{L}_{foot}}_{auxiliary} 
\end{equation}


\subsection{Applications}
To highlight the importance and potential of the proposed new video-to-motion generation method, we demonstrate how it elevates three widely used motion applications specifically in this section. 

\subsubsection{3D Motion Dataset Construction via Diffusion}

The motion dataset~\cite{ionescu2013human3,li2023finedance,mahmood2019amass,plappert2016kit,mandery2015kit,guo2020action2motion} is the cornerstone for motion-related tasks since it is essential to power up model training, feature learning and performance evaluation. However, it is time-consuming to build up a motion dataset compared with other modalities such as images or videos -- Mocap systems could be million level and some actions need to be performed by human actors in the studio. Moreover, some desirable motions like skiing, water sports or celebrity dancing are almost impossible to collect due to the difficulties of recording outdoor activities as well as recruiting professional actors. Hence, it is meaningful to design an alternative approach to construct a Motion dataset in a cost-effective and efficient manner. 

Dance data is markedly deficient in existing available motion data where Chinese classic dance is conspicuously absent from the repertoire. Hence, we first collect 52 videos from the Chinese annual dancing competition, which are all solo-dance, continuous and high-quality. Then the 2D poses are extracted by AlphaPose~\cite{fang2022alphapose} and the confidence of joints in missing frames is filled with zero. A simple interpolation and completion process is employed to revise the 2D pose sequences to be continuous and complete. Finally, we generated 3D motions using \textit{ViMo} from all these video inputs to amplify the numbers, and carefully chosen 750 clips. This dataset spans a total time length of up to 63 minutes. Our generated Chinese Classic motion dancing samples are illustrated in Fig~\ref{fig:supply_chinese_dance}. It can be seen that our ViMo can create realistic and various motions. This motion dataset expands the dance genres and could improve performance for many tasks such as music-to-motion generation, motion recognition and motion prediction. Details are dicussed in the experiment section. 


\subsubsection{Few-shot Dancing Stylization from Videos}
The existing approach to music-to-dance generation~\cite{tseng2023edge} restricts the dance generation to 1-25 predefined categories. Our model significantly overcomes this limitation by generating dances in any style using just a few reference videos. In essence, this application takes videos and music of any length as input and generates motions that adopt the pose style from the video and the beat alignment from the music. This principle can also be applied to audio-based and text-based motion generation pipelines.

We demonstrate a successful example using several figure skating videos as references. To accomplish this, we integrated zero-convolution blocks, similar to those in the basic music-to-dance diffusion baseline~\cite{tseng2023edge}. The efficiency of these zero-convolution blocks is demonstrated in experiments results. We then retrained this model using prompts from the source figure skating video. By learning the distribution of video-guided data, our model gains the ability to synthesize dance movements similar to figure skating. Examples of the generated dances are presented in Fig~\ref{fig:dance}, confirming the potential of our ViMo to extract styles and translate them between different modalities.

\subsubsection{Video-guided Motion Completion}
Motion Completion~\cite{harvey2020robust,duan2021single} plays an important role in animation and 3D-featured film production. Users only need to provide some keyframes or clips as guidance to obtain a complete movement, which significantly improves users' effort and efficiency. Based on the type of missing parts, motion completion is usually categorized into three types, in-between, in-filling and blending~\cite{duan2021single}. 

Users can perform motion in-betweening by providing a reference motion $c_{const}$ with the same range of $m \in \mathbb{R}^{S \times 151}$ and a mask $b \in\{0,1\}^{N \times 151}$, where $\boldsymbol{m}$ is all 1 's in the first and last $n$ frames and 0 everywhere else. This would result in a sequence $N$ frames long, where the first and last $n$ frames are provided by the reference motion and the rest is filled in with a plausible ``in-between" dance that smoothly connects the constraint frames.

\begin{figure*}[!h]
  \centering
  \includegraphics[width=0.99\textwidth]{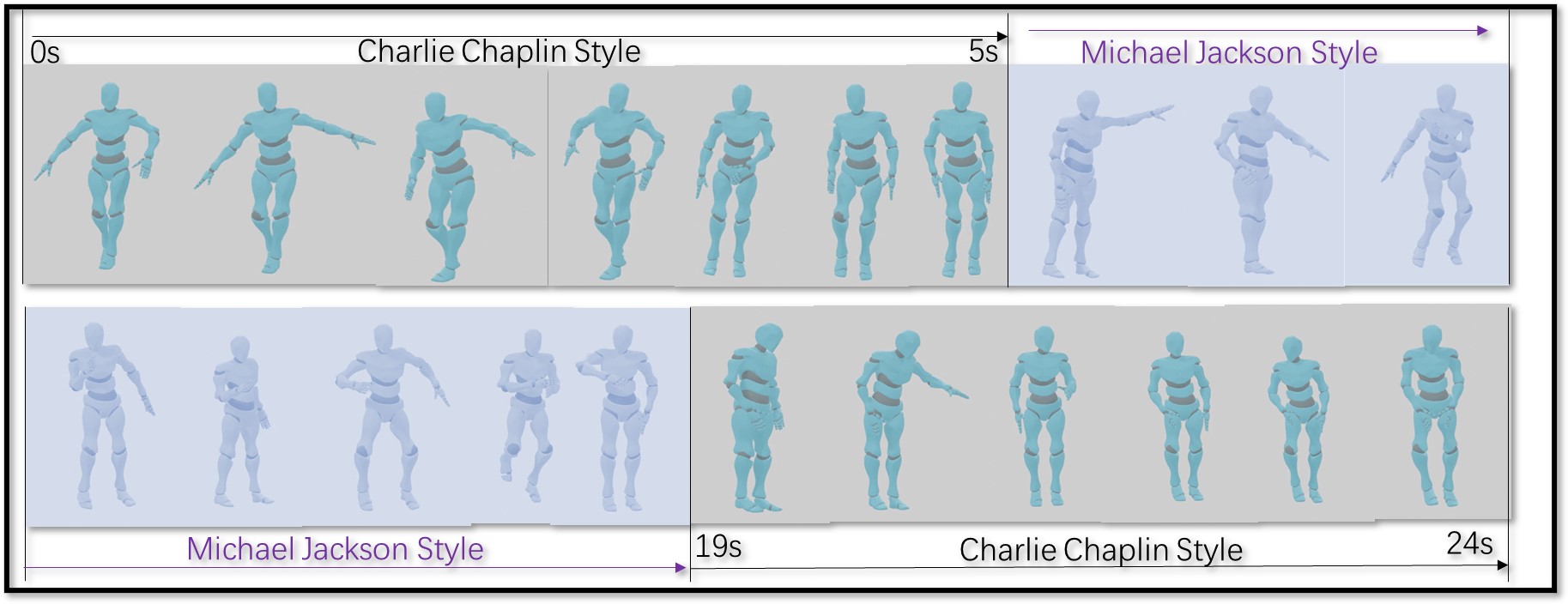}
  \caption{\textbf{Completion example} of inpainting a Charlie Chaplin style motion with a Michael Jackson dancing style clip. ViMo provides a convenient way to manipulate motion styles and generate natural and smooth motion results.}
  \label{fig:MJackson}
  \vspace{-2mm}
\end{figure*}

Inspired by the image inpainting process~\cite{lugmayr2022repaint}, the proposed diffusion model \textbf{ViMo} could be leveraged with standard masked denoising techniques to perform the motion completion. A simple example of tackling the motion in-between case, our model could expand a distribution space by the input videos to fill in the missing frames in the motion. Mathematically, given joint-wise positions indicated by a binary mask $b$, we perform the following at every denoising time step:

\begin{equation}
\hat{\mathbf{m}}_{t-1}:=\boldsymbol{b} \odot q\left(c_{const}, t-1\right)+(1-\boldsymbol{b}) \odot \hat{\mathbf{m}}_{t-1}
\end{equation}
where $\odot$ is the element-wise product, replacing the known regions with forward-diffused samples of the constraint. This technique allows editability for motion generation in referenced styles. Visualization results are demonstrated in Fig~\ref{fig:MJackson}.

\begin{figure*}[!h]
  \centering
  \includegraphics[width=0.99\textwidth]{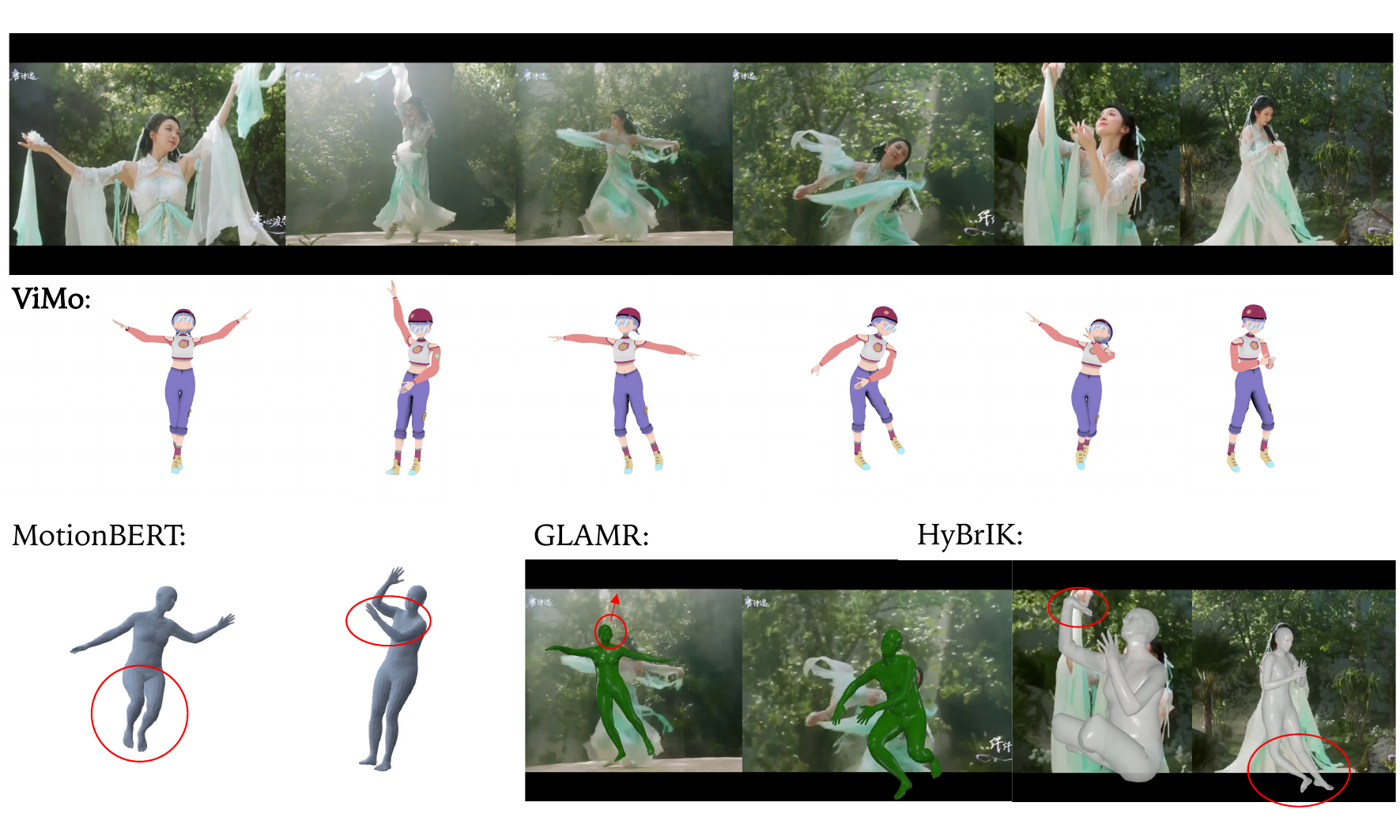}
  \caption{\textbf{Qualitative motion results with different methods.} It can be seen that proposed ViMo generate relatively more plausible motions compared with other methods.}
  \label{fig:supply_comp}
\end{figure*}

\section{Experiments}
To thoroughly assess the effectiveness of the proposed method, we have reported comprehensive evaluation metrics, alongside the outcomes of user study in the following paragraphs. 

\subsection{Configurations} 

\subsubsection{Dataset.} In this work, AIST++~\cite{li2021ai} dataset is used, which is consisting of 1,408 high-quality dance motions paired with music from 10 kinds of genres. Each 3D motion inside corresponds to 2D videos from 8 angles. Among 10 dances genres in AIST++, we used the two genres -- break and middle hip-hop for testing and the rest 8 genres for training. All training examples are cut to 5 seconds, 30 FPS. During training, the multi-view projected 2D pose sequences from different camera angles are used as input data to enable our ViMo model has the ability to reconstruct 3D skeletons from different views. 

In addition, as an important application of our ViMo, a 3D motion dataset of Chinese classic dancing via diffusion is built. This dataset contains 52 Chinese classic dancing videos together with 750 generated 3D dancing motions. This dataset would benefit tasks on generation, prediction and recognition for more study on motion and we will release the dataset. More details can be found in Section 4.2.1.


\subsubsection{Baseline and Backbone Methods.} 
We have selected two advanced state-of-the-art methods as baseline and backbone methods.

First, we employ MotionBERT~\cite{zhu2023motionbert}, which has recently drawn attention for its superior ability to reconstruct motion from videos. To conduct a thorough assessment of its proficiency in deriving 3D motions from casual videos, we retrain MotionBERT on the comprehensive AIST++ dataset to conduct a fair comparison.

Second, we incorporate EDGE~\cite{tseng2023edge}, a state-of-the-art music-to-dance conversion model employing diffusion-based methods for efficient few-shot stylization from videos, and characterized by its transformer-based diffusion framework as our backbone methods in Section 4.2.3. We test our model using a diverse range of dance videos, including Chinese classical dance, figure skating, and dances emulating Michael Jackson's style, to generate movements within these distinct categories. 

The evaluation emphasize the quality, variety, and musical synchronization of our dance generation, using the baseline methodologies as reference points. For an in-depth analysis of the metrics used, please see the following paragraph.

\subsubsection{Metrics.}
Our evaluation of the generated dances encompasses three critical dimensions: the quality of the dances, the range of motion diversity, and the synchronicity between the music rhythms and the generated movements. 

Specifically, for assessing motion quality, we employ the Frechet Inception Distances (FID)~\cite{heusel2017gans} and the Physical Foot Contact Score~\cite{tseng2023edge}. These metrics are calculated by comparing the generated dances against all motion sequences (inclusive of both training and test data) from the AIST++ dataset, focusing on kinetic features~\cite{onuma2008fmdistance} (labelled as `${FID}\_k$') and handpicked geometric features~\cite{muller2005efficient} (labelled as `${FID}\_m$’), both of which are extracted using the toolkit from~\cite{gopinath2020fairmotion}. 

For motion diversity, we compute the mean distance between features of generated movements, adhering to the method outlined in~\cite{li2021ai}, denoted as `${Div}\_k$; and `${Div}\_m$' for kinetic features and geometric features respectively. 

To quantify the alignment of music and movements, we measure the average temporal discrepancy between each musical beat and its nearest corresponding dance beat, denoting this as the Beat Align Score (BA). 

Additionally, for qualitative assessments, we conducted user studies involving 44 uniquely generated 3D motions on `Win Rate', which is the proportion of the preferred choice among candidates and indicates the winners' relevance and applicability in real-world scenarios.

\subsubsection{Implementation Details.}
We set the sequence to 5 seconds and aligned all the videos and motions with fps to 30. The target motion contains the 6D rotation extracted from the SMPL $\theta$ parameter and the global position of the root joint. The binary foot contact labels $\{0,1\}^{N \times 4}$ are pre-computed from whether foot velocities of both foot joints and toe joints along the up-axis are near zero. The sampling diffusion step T is set to be 1000. We adopt~\cite{song2020denoising} to accelerate the sampling process by predicting the posterior distribution with 50 steps for each DDIM sampling. The learning rate is set to be 1e-4, the total training epoch is 1000 and the optimizer is ADAN~\cite{xie2022adan} with a decay rate to be 0.02.

\subsection{Evaluation and Application Results.}

\subsubsection{Comparison between ViMo and MotionBERT} 
We evaluated the motion-extracting capabilities of the proposed ViMo against MotionBERT~\cite{zhu2023motionbert} on dataset AIST++~\cite{li2021ai}. 

The MotionBERT models provide two different ways to obtain human motions: 3D pose estimation (in halpe-26 skeletons) and human mesh recovery (SMPL shape and pose). We choose the human mesh recovery model to align the SMPL pose format with ViMo and simply the method to predict the SMPL pose only and retrain their model on the AIST++.
\begin{table}[tb]
\caption{\textbf{Performance comparison} with MotionBERT~\cite{zhu2023motionbert} and ViMo. Experiments are evaluated on two dance genres of AIST++~\cite{li2021ai} for testing. The subscribe $k$ and $m$ denote the metric of kinetic and manually selected geometry features respectively. The results shows the superiority of our proposed ViMo on quantitative scores and user study.} 
\footnotesize
\centering 
%
\begin{tabular}{cccccc}
\hline
Model & ${FID}\_k\downarrow$ & ${FID}\_m\downarrow$  & ${Div}\_k\uparrow$ & ${Div}\_m\uparrow$ & {Win Rate}$\uparrow$ \\
\hline
MotionBERT & $19.22$ & $9.03$ & $\textbf{5.27}$ & $5.02$ &  $23.95\%$ \\
ViMo    & $\textcolor{red}{\textbf{15.70}}$ & $\textcolor{red}{\textbf{8.25}}$ & $4.85$ & $\textcolor{red}{\textbf{5.77}}$ & $\textcolor{red}{\textbf{76.05}\%}$  \\
\hline
\label{tab:comparison} 
\vspace{-1mm}
\end{tabular}
\end{table}

As shown in Table~\ref{tab:comparison}, the left part demonstrates the quantative result on motion quality (FID) and diversity (Div). It demonstrates that our model outperforms MotionBERT~\cite{zhu2023motionbert} in terms of the FID index and Diversity Index, indicating higher quality and diversity. However, as discussed in the literature~\cite{tseng2023edge}, these indices are not always reliable. Indeed, there can be instances where perceptual realism decreases even as these indices increase. Qualitative comparison are shown in Fig~\ref{fig:supply_comp}.

To substantiate these findings, we conduct a user study and report metrics as shown in the right part of Table~\ref{tab:comparison}. In this study, we create a comprehensive questionnaire disseminated through Wenjuanxing, an online platform, to recruit participants. A total of 102 participants are enlisted to compare the performance of \textit{ViMo} and MotionBERT. We present 44 video demonstrations and asked 22 questions in the questionnaire. Participants are asked to choose which model performed better or to rate their satisfaction with the motion generation quality. The participants also involves the user study in our stylization experiments in Section 4.2.3.

Furthermore, we curate a small set of extremely challenging casual videos featuring renowned Chinese dancers and advertising videos sourced directly from the internet. These videos, characterized by continuous action but complex camera movements and editing, are representative of our target casual videos. Visual inspection of the results revealed that MotionBERT struggled with this dataset, generating unusual actions. In contrast, our model proved robust, managing to produce a dance sequence that closely resembled the original and maintained the same rhythm. It is evident that the proposed ViMo can generate smoother and more plausible motions across all view inputs. Although MotionBERT sometimes could produce more accurate results on specific frames, it exhibited a tendency towards excessive shaking and jittering. It is noteworthy that the AIST++ test dataset, which does not include many camera movements, still posed a challenge to MotionBERT's reconstruction abilities.

\subsubsection{3D Motion Dataset via Diffusion}
In collaboration with \textit{ViMo}, we successfully synthesize an extensive collection of 750 Chinese classic dance motion sequences, as illustrated in Fig~\ref{fig:supply_chinese_dance}. These sequences are remarkably faithful to the original dance videos, capturing the essence and fluidity of Classic dance. We divide each extended video into segments of 5 seconds, creating concise motion clips. This segmentation significantly streamline the training process for our music-to-motion synthesis tasks. To assess the practicality of this dataset, we utilize it as a foundational training set for a music-to-dance generation task. The outcomes were impressive, showcasing an increased variety in dance and the generation of authentic Chinese classic dance styles that seamlessly synchronize with music of varying lengths. Remarkably, \textit{ViMo} possesses the potential to expand this dataset by 5 or even 50 times, paving the way for large-scale data generation from limited video resources. This methodology holds promise for application across diverse video categories.

\subsubsection{Few-shot Dancing Stylization from Videos}

The task few-shot dancing stylization focusing on generate 'music-aligned' motion in this style given these limited videos. Our approach involve the analysis of merely two figure skating videos, each sourced from Olympic competitions.  The dynamic nature of figure skating, with its unique movement patterns distinct from Chinese classic dance forms, presented a challenge. These patterns often involve repetitive and iconic sequences. Due to the sport's dynamism, the camera keeps motion to make the skater remained the focal point. 

To obtain this task, we first build a small dataset with figure skating data samples paired with music from the videos as discussed in the previous Section. Next, the backbone method EDGE~\cite{tseng2023edge}, which is the state-of-the-art music-to-dance diffusion-based method, is adopted to align the generated motion given arbitrary music. We find that directly finetuning the EDGE model with new data samples works basically, except for the learned style influenced by EDGE training data styles. This influence makes the generated motions partially consist of the street dances style in AIST++. 

To improve this, we design a zero convolution module to make the generated motions focus more on the distribution in the expected styles in the few-shot videos while keeping benefits in the well-trained parameters from EDGE checkpoints. Specifically, the zero convolution module consists of new trainable networks with the same structure and a zero convolution network attached and concatenated with the original freezing models~\cite{zhang2023adding}. The newly attached structure is initialized by the original checkpoints but can be trained to learn the new style distributions while the freezing parts provide the learned music-to-dance ability in the pre-trained backbone checkpoints.

To demonstrate the merits of the zero convolution design for our few-shot dancing stylization, we test our models using a diverse range of dance videos, including Chinese classical dance, figure skating, and dances emulating Michael Jackson's style (labelled as Type-1,2,3 respectively), to generate movements within these distinct categories as shown in Table~\ref{tab:example}. It can be seen that our proposed design labelled by `ZeroConv' shows superiority especially for the `Win Rate' in the user study.




\begin{table}[tb]

\centering
\caption{\textbf{Evaluation} on motion generation  metrics for the proposed design. The `ZeroConv' not only approximately outperforms the directly fintuning in all the quantitatively scores in BA, PFC and diversity, but also notably preferred for the \textcolor{blue}{`Win Rate'} in our user study. Note that each type is denoted in a distinct style category.}
\footnotesize
\begin{tabular}{llllll}
\hline
\hline
Type-1  & BA $\uparrow$ & PFC $\downarrow$ & $Div\_k \uparrow$ & $Div\_m \uparrow$ & Win Rate $\uparrow$ \\
\hline
ViMo+ZeroConv &  $\textbf{0.243}$   &  $\textbf{1.49}$ &  $4.32$   & $4.59$       &   $\textcolor{red}{\textbf{56.27}}\%$    \\
ViMo+finetune       &  $0.239$      &  $1.59$      &  $\textbf{4.42}$      &  $\textbf{4.80}$      &    $43.73\%$     \\
\hline
Type-2  & BA $\uparrow$ & PFC $\downarrow$ & $Div\_k \uparrow$ & $Div\_m \uparrow$ & Win Rate $\uparrow$ \\
\hline
ViMo+ZeroConv &  $\textbf{0.236}$   &  $1.50$      &  $\textbf{4.51}$       &   $4.32$     &   $\textcolor{red}{\textbf{65.89}}\%$       \\
ViMo+finetune       &  $0.226$   &  $\textbf{1.44}$    &   $4.20$     &   $\textbf{4.51}$     &    $34.11\%$  \\
\hline
Type-3 & BA $\uparrow$ & PFC $\downarrow$ & $Div\_k \uparrow$ & $Div\_m \uparrow$ & Win Rate $\uparrow$ \\
\hline
ViMo+ZeroConv &   $0.239$   &  $\textbf{1.47}$      &  $\textbf{4.78}$      &   $\textbf{4.89}$     &    $\textcolor{red}{\textbf{57.65}}\%$   \\
ViMo+finetune       &   $\textbf{0.244}$   &  $1.52$      &    $4.70$   &   $4.73$     &   $43.35\%$ \\
\hline
\hline
\label{tab:example}
\end{tabular}
\vspace{-4mm}
\end{table}

\subsubsection{Video-guided Motion Completion}
We aim to showcase our model's proficiency in accomplishing tasks such as completing interrupted motions, motion in-between, and seamlessly merging movements. We selected a small collection of casual videos in specific styles, such as depicting characters like Michael Jackson and Charlie Chaplin, to demonstrate this capability. Given expected condition control in different timestamp, ViMo can generate natural motions and smoothly switching between these styles . The results are shown in Fig~\ref{fig:MJackson}. The reconstructed motions not only include significant characteristics from the previous style but also show a high degree of fidelity in target motion style. These results illustrate ViMo's ability to control motions in expected style, offering users an intuitive approach to motion editing and semantic control.

\section{Conclusion}
In this paper, we innovate a new question to generate multiple possible 3D motions from the casual video which films complicated actions with varied camera positions and complex editing. To address this problem, we reveal a novel transformer-diffusion-based framework which could leverage the 2D pose sequences to similar 3D motions without explicitly estimating the camera angles. Visualization comparison results show that our method could outperform the existing work dramatically. Experimental results show that our method could empower three crucial motion applications, thus offering remarkable value for academics and industry. 
Our work is the first attempt toward this goal and we believe it will open new opportunities and encourage future research to exploit video for more complex motion generation, achieving more semantic, scene-aware and multi-person stages. Striking breakthroughs and tremendous benefits could be expected in the future with the exploration of this direction.

\bibliographystyle{splncs04}
\bibliography{ref}

\end{document}